\documentclass[runningheads]{llncs}
\usepackage[T1]{fontenc}
\usepackage{graphicx}
\usepackage{rotating}
\usepackage{tabularx}
\usepackage{makecell}
\usepackage{multirow}
\usepackage{enumitem}
\usepackage{amsmath}
\usepackage{caption}
\usepackage{orcidlink}
\usepackage{amsfonts} 
\usepackage{cancel}
\usepackage{pstricks}
\usepackage{mathabx}
\usepackage{tikz}
\begin{document}

\title{Resource-efficient Medical Image Analysis with Self-adapting Forward-Forward Networks}

\titlerunning{SaFF-Net}
%

\author{Johanna P. M\"uller \inst{1}\orcidlink{0000-0001-8636-7986} \and
Bernhard Kainz\inst{1,2}\orcidlink{0000-0002-7813-5023}}

\authorrunning{ M\"uller and Kainz}

\institute{
Friedrich–Alexander University Erlangen–N\"urnberg, DE 
\email{johanna.paula.mueller@fau.de} \and
Imperial College London, SW7 2AZ, London, UK
}
\maketitle              
\begin{abstract}

We introduce a fast Self-adapting Forward-Forward Network (\emph{SaFF-Net}) for medical imaging analysis, mitigating power consumption and resource limitations, which currently primarily stem from the prevalent reliance on back-propagation for model training and fine-tuning. 
Building upon the recently proposed Forward-Forward Algorithm (FFA), we introduce the Convolutional Forward-Forward Algorithm (CFFA), a parameter-efficient reformulation that is suitable for advanced image analysis and overcomes the speed and generalisation constraints of the original FFA. To address hyper-parameter sensitivity of FFAs we are also introducing a self-adapting framework \emph{SaFF-Net} fine-tuning parameters during warmup and training in parallel.
Our approach enables more effective model training and eliminates the previously essential requirement for an arbitrarily chosen Goodness function in FFA.
We evaluate our approach on several benchmarking datasets in comparison with standard Back-Propagation (BP) neural networks showing that FFA-based networks with notably fewer parameters and function evaluations can compete with standard models, especially, in one-shot scenarios and large batch sizes. 

\end{abstract}

\section{Introduction}

Efficient neural networks are becoming tremendously important for effective impact on real clinical workflows. This is particularly important as the need to fine-tune models on compute and power-constrained clinical devices becomes increasingly prevalent. Additionally, the concentration of resources in only a few specialised institutions highlights the importance of creating models that can operate efficiently in various low-resource settings, ensuring widespread access to advanced healthcare technologies. Moreover, the lack of computational infrastructure in areas that could benefit the most from population-fine-tuned deep learning models due to the scarcity of experts underscores the urgency of developing compute and energy-efficient neural networks that can transcend such limitations and make a meaningful impact on global healthcare disparities. This paper addresses the pressing need to develop neural architectures that not only perform well but also operate with optimal energy consumption during training and application. To this end, we introduce a fast self-adapting Forward-Forward Network without using Back-Propagation (BP) for medical imaging analysis to mitigate power consumption and enable training if computational resources are limited, contributing to the larger discourse on responsible AI development. 

Based on preliminary work on the Forward-Forward Algorithm (FFA)~\cite{hinton2022forward}, we know that FFAs can be slightly slower than BP and seem not to generalize as well on several simplified example datasets. In this paper, we address the remaining issues of FFAs and provide comprehensive insights into the advancement of FFAs. To ensure reproducibility and adoption, we also suggest and share an automatically self-configuring FFA framework for medical imaging analysis considering the limitations of energy and resources. 

\noindent\textbf{Contributions. }
    \textbf{(1)} We present the Convolutional Forward-Forward Algorithm as an extension to the Forward-Forward Algorithm by~\cite{hinton2022forward}. 
    \textbf{(2)} We show that the (Convolutional) Forward-Forward Algorithm can be utilised fully without backpropagation for classification and pretraining of Multi-Layer-Perceptrons and other Convolutional Neural Networks even without the use of domain-limited goodness functions.  
    \textbf{(3)} We propose the backpropagation-free self-adaptive Forward-Forward Framework (SaFF-Net) and show its advantages, evaluating the numbers of function evaluations and parameters, and its general performance.

\noindent\textbf{Related Work.} 
The Forward-Forward Algorithm (FFA)~\cite{hinton2022forward} diverges from the traditional Backpropagation (BP) algorithm similar to Evolutionary Algorithms ~\cite{stanley2002evolving,salimans2017evolution}, Predictive Coding~\cite{spratling2017review}, Hebbian learning~\cite{lagani2022comparing}, Feedback Alignment~\cite{lillicrap2016random} and Reservoir Computing~\cite{bianchi2020reservoir}. These alternatives to BP exhibit challenges such as elevated computational demands, sensitivity to noise, a lack of explainability, and difficulties in training and global optimisation.

\cite{hinton2022forward,lorberbom2024layer,gandhi2023extending,yang2023theory,hopwood2023one,pardientropy} explore the FFA, offering insights into its behaviour, strengths, and limitations. These works also propose incremental extensions for enhancing FFA performance. From a biological perspective, FFA demonstrates potential superiority over BP algorithms \cite{tosato2023emergent,gandhi2023extending}. \cite{hinton2022forward} presents FFA as a biological model, notably for modelling sleep deprivation \cite{hinton2022forward,licua2023sleep}.

Eliminating the need for computationally intensive BP, the transition to analogue hardware, such as microcontroller architectures \cite{de2023mu}, and the application of FFA in Spiking Neural Networks \cite{ororbia2023predictive,ororbia2023learning}, appear promising.  
The integration of FFA within Recurrent Neural Networks (RNNs) has  been explored \cite{ororbia2023predictive} and the application of FFA in Graph Neural Networks (GNNs) has gained attention for handling complex relationships in graph-structured data efficiently \cite{paliotta2023graph}. 
For Self-Supervised Learning (SSL),~\cite{brenig2023study} discusses data augmentation techniques in training FFAs. They propose self-supervised FFAs as a superior approach compared to the unsupervised approach presented by \cite{hinton2022forward}.

\cite{ahamed2023ffcl} explores FFA pretraining, presenting a framework with two stages of local and global contrastive representation learning, and subsequent BP-based image classification with binary-cross-entropy. They replace the MLP with a ResNet18 architecture. Other works also investigate the synergy between FFA and BP techniques, aiming to leverage the strengths of both approaches for enhanced learning and overall energy efficiency \cite{giampaolo2023investigating,brenig2023study}.

\noindent\textbf{FFA Limitations.}
The FFAs goodness function is important for layer-wise training in a contrastive learning setting. The positive pass adjusts weights on real data to increase goodness, while the negative pass handles counter-examples, adjusting weights to decrease goodness. However, this idea poses three major problems. (1) It requires an additional dataset-specific hyperparameter, $\theta$, for thresholding. This hyperparameter is dataset-specific and needs independent optimisation. (2) For prediction, assessing goodness for each possible label limits applications to classification. A mitigation is presented in \cite{ororbia2023predictive}, using FFAs in Recurrent Neural Networks for generative and reconstructive tasks. There, the goodness function only regulates neuron activity, resulting in the last layer's output containing no spatial information; only the sum of goodness over all layers determines the result. 
(3) Based on preliminary experiments and results from \cite{lorberbom2024layer}, adding more than two layers does not positively contribute to FFA performance. The consequent dependence on Multi-Layer Perceptron (MLP) architectures and heuristic designs for the goodness function limits FFA significantly for practical applications. 
To the best of our knowledge, only  \cite{ahamed2023ffcl} pretrains a CNN, specifically a ResNet18~\cite{he2016deep} backbone, for classification. In parallel to our work another convolutional approach was proposed by \cite{papachristodoulou2023convolutional} including a complex channel-wise competitive learning strategy.
However, the primary constraint of these methods lies in their extensive need for hyperparameter tuning, encompassing a wide array of parameters such as threshold settings, the iteration count per layer, the total number of epochs, the architecture depth (number of layers), learning rate, batch size, the choice of loss and activation functions. Specifically for convolutional feature extraction, this extends to configuring the number of input and output channels, stride lengths, padding types, and kernel dimensions.

\section{Method}

\begin{figure}[h]
\centering
\includegraphics[width=0.95\textwidth]{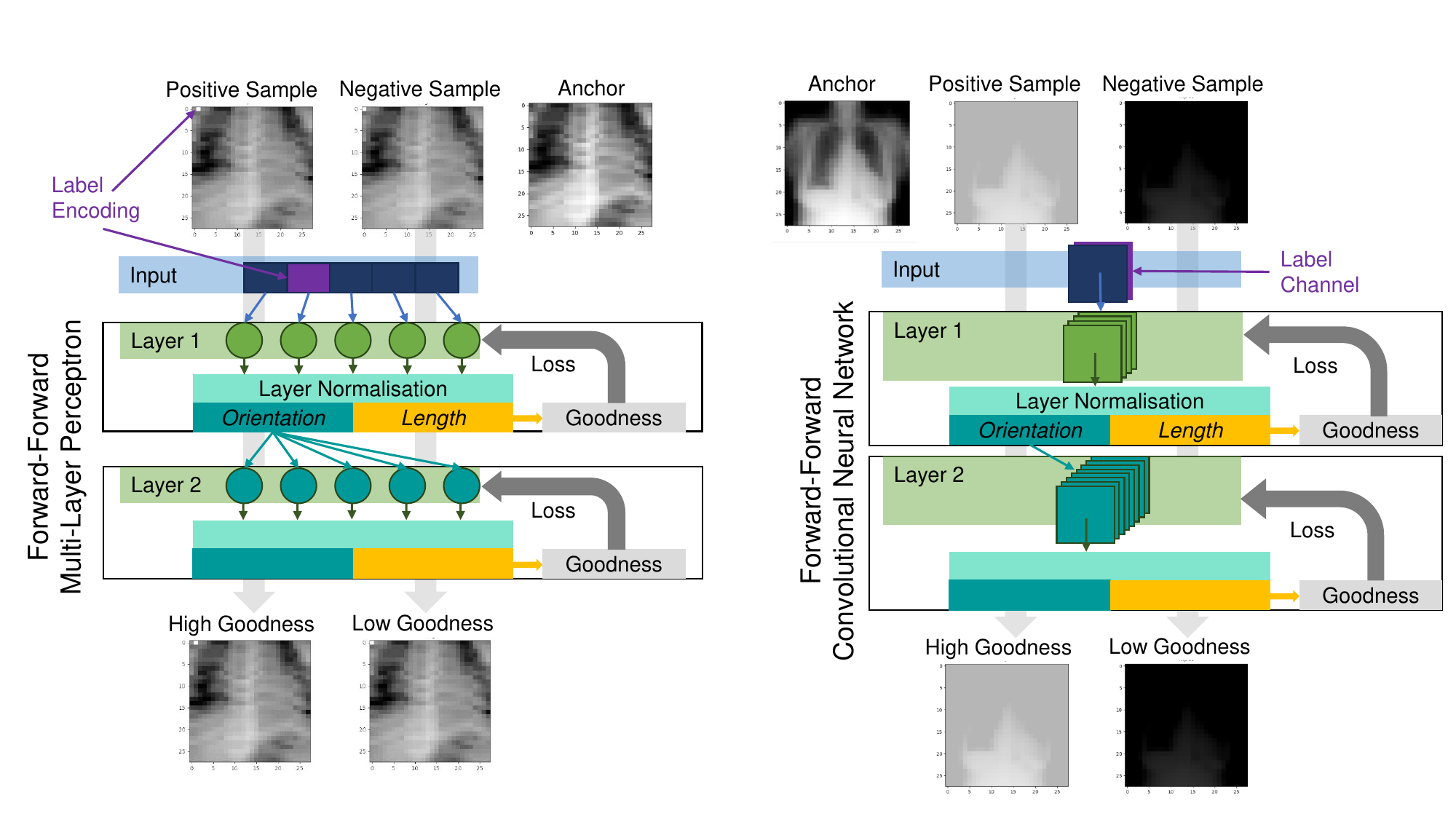}
\caption{The Forward-Forward Multi-Layer Perceptron (left), the Forward-Forward Convolutional Neural Network (right). The networks are optimised layer-wise. The positive and negative samples are fed into the first layer and via layer normalisation, we obtain orientation and length of the activation vector. The orientation is forwarded to the next layer as its input. The length is used for the computation of the goodness. Each layer is optimised so that positive samples have high goodness (> threshold) and negative samples have low goodness (< threshold). For inference, the sum of the goodness of all layers, excluding the first layer, needs to be determined for every possible label.}
\label{fig:ff}
\end{figure}

The Forward-Forward Algorithm~\cite{hinton2022forward} (Fig. \ref{fig:ff}, left) is a multi-layer learning approach influenced by Boltzmann machines~\cite{hinton1986learning} and Noise Contrastive Estimation~\cite{gutmann2010noise}. Unlike traditional BP, it employs two forward passes (instead of forward and backward) operating similarly but on different data with opposing objectives. The positive pass operates on real data $x^{+}$ and adjusts the weights to increase the goodness. The negative pass operates on counter-examples $x^{-}$ and adjusts the weights to decrease the goodness. The goodness function 

\begin{equation}\label{goodness}
g(x) = \sum_{j} y_{j}(x^{+/-})^{2} -\theta
\end{equation}

can be defined as the (negative) sum of the squared neural activities of $y(x)$, where $y_j$ is the output of hidden unit $j$ and $x^{+/-}$ is the input.
The probability $p(x)$ that an input vector $x$ is real (true) is given by $p(x) = \sigma(g(x))$, where $\sigma$ is the logistic function. However, if the activities of the first hidden layer are used as input for the second hidden layer, distinguishing between positive and negative data becomes straightforward. This can be achieved by simply measuring the length of the activity vector from the first hidden layer, without the need to learn any new features. To mitigate this issue, the FFA employs a normalisation on the length of the hidden vector before feeding it as input to the subsequent layer. It follows, that only the length of $y^{+/-}$ is used for goodness and the orientation derived from the (layer) normalisation is forwarded to the next hidden layer. \cite{hinton2022forward} 

\noindent\textbf{Convolutional Forward-Forward Algorithm.}
We propose the Convolutional Forward-Forward Algorithm (CFFA) and extend the application of the FFA to Convolutional Neural Networks (CNNs), see Fig. \ref{fig:ff}, right. Given the two-dimensional convolution in an arbitrary convolutional layer

\begin{equation}
    y_{i,j} = \alpha\left(\sum_{m=1}^{M}\sum_{n=1}^{N} x_{m,n} \cdot w_{i,j,m,n} + b_{i,j}\right),
\end{equation}

where $y_{i,j}$ is the output feature at position $(i, j)$, $x_{m,n}$ is the input feature at position $(m, n)$, $w_{i,j,m,n}$ is the weight at position $(i-m+\lfloor \frac{K}{2} \rfloor, j-n+\lfloor \frac{K}{2} \rfloor)$ in the kernel with size $K$, $b_{i,j}$ is the bias term at position $(i, j)$, and $\alpha(\cdot)$ is the activation function. The same applies to three-dimensional convolutions.
Consequently, we can also use the goodness and the probability function for a given hidden layer with input $x \in \mathbb{R}^{MxN}$. Therefore, we can take the length of $y^{+/-}_{i,j}$ for goodness and the orientation derived from the (layer) normalisation can be forwarded to the next hidden layer, maintaining spatial dependencies. At this point, we want to emphasise that convolutional layers in CFFAs do not impede potential on-chip training for tiny devices, see exemplarily \cite{wang2016low,yamazaki2022spiking}. 

\noindent\textbf{Objective.}
Given the goodness function $g(x)$ (Eq. \ref{goodness}), we can formulate the optimisation problem with an arbitrary contrastive loss function $F(x^{+},x^{-})$

\begin{equation}
    \min_{x} F(x) = \min_{x} \frac{1}{B} \sum_{i=1}^{B} \log(G(x_i))
\end{equation}

where $B$ is the dimension of the input batch and goodness matrix $G_i = (- g(x_{i}^{+}) \mathbin\Vert g(x_{i}^{-}))$ with concatenation operator $(\mathbin\Vert)$. 
The objective function can be easily replaced by any contrastive loss function for contrastive pretraining.\\

\noindent\textbf{Self-adapting SaFF-Net.} 
We propose an FFA-based framework (see Appendix for an overview). 
Related to energy-based neural networks as the FFA itself, it is able to minimise the energy for training and necessary for inference additionally leading to a resource-efficient model optimisation framework.
Initially, we fine-tuned hyperparameters such as learning rate, batch size, number of layers, and sizes of filters and weights for task-specific optimization, settling on a configuration that delivered optimal average performance across all datasets. 
In a warm-up stage, these configuration sets can be pre-evaluated and selected based on preceding statistical tests. We use Early Stopping, where the training process is halted based on the model's performance on a validation set. This not only prevents overfitting but also expedites convergence by recognizing when further iterations yield diminishing returns. We apply Early Stopping not only on layer iterations and epochs but also on the number of layers, hence, restricting the depth of the network. We also include network pruning~\cite{blalock2020state} and incorporate Peer-Normalisation and Batch-Normalisation Loss as an additional loss term during training.
Additionally, we introduce a trainable threshold parameter $\theta$ for the goodness function for reducing the number of hyperparameters. For training, each dataset is Z-score normalised within the framework by default but other normalisations or augmentation functions can be easily added through customised experiment files. Finally, SaFF dynamically selects an optimised architecture during the warm-up phase based on prior statistical analysis of dataset characteristics and during training based on model performance.

\noindent\textbf{Implementation.} We use Python and PyTorch. All models were trained on NVIDIA A100 $80$GB. The average training time for a model with maximum batch size depends on the number of layer iterations and/or epochs and is $< 30$ min for one-shot learning and $< 2$ h for others. 

\section{Evaluation}

\noindent\textbf{Classification.}
We can use FFA and CFFA for the classification of closed-set problems. We produce positive and negative samples through label encodings in the image data. For a $n$-class problem, we take, \emph{e.g.}, the first $n$ pixels for the label encoding, setting position $i=0...n$ with the maximum intensity value according to the true label for positive samples. For negative samples, we set, \emph{e.g.}, a random label encoding position with the maximum intensity value in an unsupervised setting. We investigate different self- and semi-supervised tasks to show the generalizability of our method. We evaluate the performance on an independent test set with Accuracy, Area Under the Receiver Operating characteristic (AUC) and mean Average Precision (mAP).

\noindent\textbf{Pretraining.}
We can contrastively pretrain in unsupervised, self-supervised and supervised settings. In the pretraining phase, the focus lies on training solely the encoder, subsequently freezing it and learning the decoder with the input of the encoder. This approach ensures the preservation of the encoder's weights, shielding them from any alteration induced by BP during the subsequent downstream task. We test pretraining with FFA, CFFA and ResCFFA on three downstream tasks, Classification, Reconstruction, and Segmentation. For segmentation, we evaluate the performance on an independent test set with Accuracy and Intersection over Union (IoU). Mean Structural Similarity Index (MSSI) and Peak Signal-to-Noise Ratio (PSNR) are applied to evaluate reconstruction. 

\noindent\textbf{Datasets.}
For baseline comparisons we evaluate  on \href{http://yann.lecun.com/exdb/mnist/}{MNIST}~\cite{deng2012mnist} and \href{https://medmnist.com/}{MedMNIST}~\cite{medmnistv2}. 
To test real-world scalability, we use \href{https://vindr.ai/datasets/cxr}{VinDr-CXR}~\cite{nguyen2020vindrcxr}, part of the \href{https://vindr.ai/datasets}{VinDr}. VinDr consists of 100,000  thoracic chest X-ray images and annotations provided by 17 radiologists. We focus on the binary classification, healthy and unhealthy, and the segmentation of Cardiomegaly. 

\section{Results}

\begin{table}[t] 
\caption{Ablation study for \emph{SaFF-Net} self-configuration  on MNIST and MedMNIST (Supervised). The FFA Baseline  consists of 4 layers with 1000 neurons each. \emph{One-shot} means the network is trained on one single-batch. Networks are trained max. 10 epochs with 1,000 layer iterations. The optimised number of parameters for \emph{SaFF-Net} is given in the last row.  number of function evaluations (NFE) - Forward and Backward passes $\times1e^{3}$, ES - Early Stopping}
\resizebox{1.0\columnwidth}{!}{
\begin{tabular}{lcccccccccccc}
 & \multicolumn{6}{c}{MNIST} & \multicolumn{6}{c}{MedMNIST Pneumonia}\\
  \cmidrule(lr){2-7} \cmidrule(lr){8-13}
 &  \multicolumn{3}{c}{FFA (\emph{One-shot})}  &  \multicolumn{3}{c}{FFA}  &  \multicolumn{3}{c}{FFA (\emph{One-shot})} &  \multicolumn{3}{c}{FFA}\\

 & ACC$\uparrow$ & AUC$\uparrow$ & NFE$\downarrow$ &  ACC$\uparrow$ & AUC$\uparrow$ & NFE$\downarrow$  &  ACC$\uparrow$ & AUC$\uparrow$ & NFE$\downarrow$ &  ACC$\uparrow$ & AUC$\uparrow$ & NFE$\downarrow$  \\
  \cmidrule(lr){2-4} \cmidrule(lr){5-7} \cmidrule(lr){8-10} \cmidrule(lr){11-13} 
 Baseline & 0.24 & 0.83 & \underline{8.0} & 0.95 & \underline{0.99} & 80.0 & 0.84 & 0.84 & \underline{8.0} & 0.87 & 0.90 & 80.0\\
 \midrule
 ES Iterations & \underline{0.93} & \underline{0.99} & \textbf{7.3} & 0.11 & 0.55 & \textbf{9.3} & \textbf{0.87} & \textbf{0.91} & \textbf{1.5} & 0.83 & 0.89 & \textbf{2.0}\\
 ES Layers & 0.24 & 0.83 & \underline{8.0} & \underline{0.96} & \textbf{1.00} & 60.0 & 0.84 & 0.85 & \underline{8.0} & \underline{0.87} & 0.90 & 60.0\\
 ES Epoch & - & - & - & 0.95 & \textbf{1.00} & 80.0 &  - & - & -  & 0.84 & 0.89 & 56.0 \\
 \midrule
 Pruning (0.4) & 0.22 & 0.82 & \underline{8.0} & 0.85 & \underline{0.99} & 80.0 & \underline{0.85} & 0.84 & \underline{8.0} & 0.82 & 0.83 & 80.0\\
 w/ Retraining & 0.22 & 0.82 & 8.8 &  0.85 & \underline{0.99} & 80.8  & \underline{0.85} & 0.84 & 8.8 & 0.81 & 0.83 & 80.8\\
 \midrule
 Warm-Up & 0.01 & 0.42 & 32.0 & 0.64 & 0.93 & 104.0 & 0.80 & \underline{0.86} & 32.0 & 0.83 & \underline{0.91} & 80.0 \\
 Train. Thresh.  & 0.84 & 0.97 & \underline{8.0} &  \textbf{0.97} & \textbf{1.00} & 80.0  & 0.83 & 0.72 & \underline{8.0} & \textbf{0.89} & \textbf{0.94} & 80.0\\ 
 Peer-Norm  & 0.23 & 0.83 & \underline{8.0} & 0.95 & \textbf{1.00} & 80.0 & 0.83 & \underline{0.86} & \underline{8.0} & 0.85 & 0.89 & 80.0\\ 
 \midrule
\emph{SaFF-Net} & \textbf{0.97} & \textbf{1.00} & 29.9 &  \textbf{0.97} & \textbf{1.00} & \underline{29.9} & \textbf{0.87} & \textbf{0.91} & 15.8 & \underline{0.87} & \underline{0.91}  & \underline{15.7} \\
 \cmidrule(lr){2-4} \cmidrule(lr){5-7} \cmidrule(lr){8-10} \cmidrule(lr){11-13}
 \#Par. &  \multicolumn{3}{c}{1,144,004} & \multicolumn{3}{c}{893,503}  & \multicolumn{3}{c}{3,788,004}  & \multicolumn{3}{c}{2,787,003} \\
 \label{tab:abl-ffa}
\end{tabular}
}
\vspace{-0.7cm}
\end{table}

\begin{figure}[t]
\centering
\includegraphics[width=\textwidth]{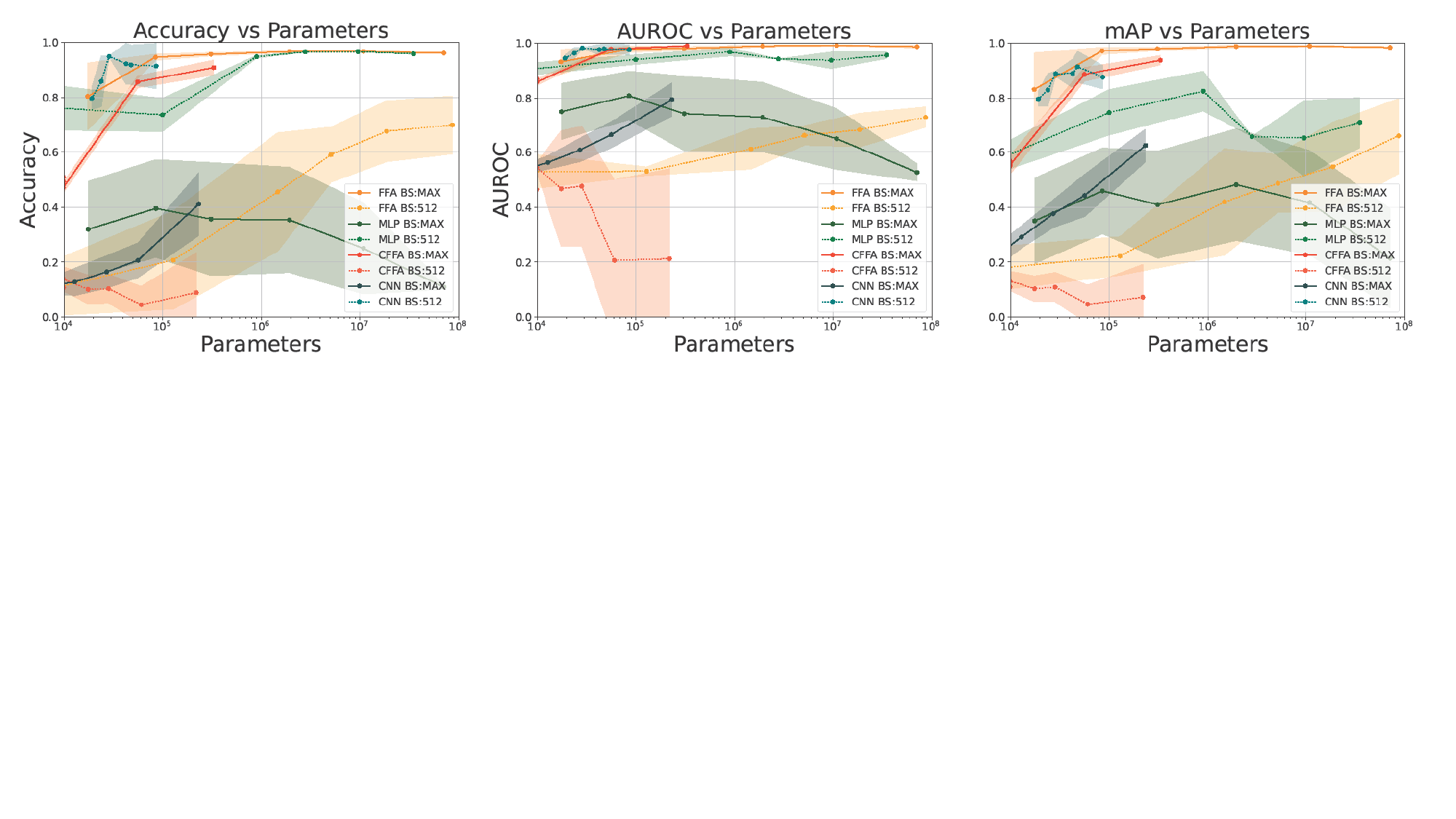}
\vspace{-4.5cm}
\caption{Classification on MNIST. ACC - Accuracy, AUC - Area Under the Receiver Operating characteristic, mAP - Mean Average Precision vs. Number of Parameters Comparison for MLP and FFA (top) and CNN and CFFA (bottom) with maximum batch size. Ours in orange.}
\label{fig:efficiency}
\end{figure}

\begin{table}[t]
\caption{Classification performance for Semi-supervised (SemiSL), self-supervised (SSL), and average from established methods for fully supervised (SL). SaFF-Net with FFA is used for 1D (flatten) input data and CFFA is used for 2D/3D input data. 
}
\resizebox{1.0\columnwidth}{!}{
\begin{tabular}{lllcccccccccccccc}
& & & & & \multicolumn{8}{c}{MedMNIST} & \multicolumn{2}{c}{VinDr-CXR} \\
 \cmidrule(lr){6-13} \cmidrule(lr){14-15}
 & &  & \multicolumn{2}{c}{\scriptsize{MNIST}} & \multicolumn{2}{c}{\scriptsize{Pneumonia}} & \multicolumn{2}{c}{\scriptsize{OCT}} & \multicolumn{2}{c}{\scriptsize{OrganA}} & \multicolumn{2}{c}{\scriptsize{Synapse3D}} & \multicolumn{2}{c}{\scriptsize{Binary}} \\
 & & \#par. & ACC & AUC &  ACC & AUC &  ACC & AUC &  ACC & AUC &  ACC & AUC &  ACC & AUC \\
 \multirow{1}{*}{\rotatebox{90}{\makecell[l]{SL}}}&\cite{NECO_a_00052,2304-10054,medmnistv1} $\diameter$ && 89.75 & 95.9  & 84.27 & 91.27 & 68.93 & 77.63 & 78.67 & 91.27 & 61.07 & 58.53 & - & -  \\
 \midrule
  \multirow{3}{*}{\rotatebox{90}{\makecell[l]{SemiSL}}}  &C-Mixer\cite{2304-10054}& 6m & - & - &  56.6 & 79.9 & \textbf{72.8} & \underline{69.9} & - & - & - & - & - & - \\
   & FFA (OS)(\emph{Ours})& 2.8m &91.0 & \underline{99.0} & 51.0 & 52.6 & 19.2 & 44.2 & 33.3 & 72.1 & \textbf{73.0} & \textbf{74.8} & 75.7 & \underline{81.5} \\
   & CFFA (OS)(\emph{Ours}) & 190k & \underline{92.5} & \textbf{99.2} & \textbf{86.2} & \textbf{93.1} & \underline{70.1} & \textbf{89.5} & \underline{34.9} & \textbf{80.2} & \underline{63.2} & \underline{67.2} & 78.0 & 80.1\\
 \midrule
  \multirow{2}{*}{\rotatebox{90}{\makecell[l]{SSL}}} & FFA (\emph{Ours})& 2.8m & \textbf{92.9} & 98.8 & 62.5 & 58.7 & 24.5& 51.7 & 23.0 & 49.0  & 27.0 & 27.6 & \underline{78.6} & 81.3\\
   & CFFA (\emph{Ours}) & 190k & 92.1 & 91.2 & \underline{85.7} & \underline{84.9} & 44.5 & 63.7 & \textbf{58.2} & \underline{79.2} & 33.7 & 64.0 & \textbf{99.2} & \textbf{99.6}&\\
 \label{tab:sota}
\end{tabular}
}
\vspace{-0.5cm}
\end{table}

\noindent\textbf{Classification.}
Prior experiments show that the optimised threshold value of $\theta$ is correlated with the test performance. We tested this correlation hypothesis and rejected the null hypotheses for all three metrics (p-values for Accuracy: 0.061, AUC: 0.051, mAP: 0.065). 
This significance only exists for threshold $\theta$ of the first layer if the first two layers were trained. This enables us to use $\theta$ for selecting the number of neurons per layer during warm-up. For evaluating the influence of all possible components in our framework, we studied ablations for FFA~\ref{tab:abl-ffa} and CFFA (Suppl. Material, Tab. 4), for MNIST and an exemplary MedMNIST dataset. Our largest FFA model configuration consists of $3,788,004$ parameters which is only necessary for one-shot learning. Based on the ablation study the components with the highest impact are Early Stopping for layer iterations and the trainable threshold parameter in FFA models, especially, for epoch-wise learning. For one-shot learning, the trainable parameter had less influence on the performance. For CFFA models, Early Stopping for layer iterations was less conducive, on the contrary, the trainable threshold parameter increased the performance.

All CFFA models have the same size of $194,432$ parameters similar to the ResCFFA model with $148,933$ parameters. The ResCFFA is built of 5 2D-CFFA (64-channel) layers with one residual connection between the first hidden layer and the fourth. We test against three benchmark models an MLP with $2,797,010$ (increases for multi-channel input) parameters (3 hidden layers à 1000 neurons plus 1 fully connected output layer), a CNN with $1,554,954$ parameters (4 hidden layers with [64,128,256,512] features plus 1 fully connected output layer; it performed worse with same configuration as CFFA) and ResNet18 with $11,183,694$ parameters. We present an extensive table comparing networks with and without BP in Suppl. Material, Tab. 5.
The comparison between FFA and MLP models highlights that we can achieve in the mean for the MedMNIST dataset an improvement by 22 \% in accuracy with \emph{SaFF-Net}. For the 3D MedMNIST data, the MLP performed slightly better (7\%). In comparing CFFA models with CNNs, our results show an overall performance increase when using CFFA for MedMNIST (10\%), for 3D the CNN performed 37 \% better. We compare the results of our models with comparable SOTA methods and show the performance for semi-supervised learning ($10 \%$ labelled) and self-supervised pretraining for SaFF-Net in Tab.~\ref{tab:sota}.

\noindent\textbf{Pretraining.}
We contrastively pre-train \emph{SaFF-Net} models exemplary with the Triplet Margin Loss~\cite{balntas2016learning}. We investigated the different classification, reconstruction, and segmentation heads for three datasets, see Tab.~\ref{tab:pre}. For segmentation, we test on the Cardiomegaly subset of VinDr-CXR.
\vspace{-0.7cm}
\begin{table}[h!]
\caption{Pretraining \emph{SaFF-Net} with FFA, CFFA and ResCFFA. ACC - Accuracy, AUC - Area Under the Receiver Operating characteristic, AP - Mean Average Precision, SSIM - Mean Structural Similarity Index, PSNR - Peak Signal-to-Noise Ratio, IoU - Intersection over union.$*$ with Triplet Margin Loss}
\resizebox{1.0\columnwidth}{!}{%
\begin{tabular}{lccccccccccc}
&& \multicolumn{4}{c}{Classification$*$} & \multicolumn{4}{c}{Reconstruction} & \multicolumn{2}{c}{Segmentation} \\
\cmidrule(lr){3-6}\cmidrule(lr){7-10}\cmidrule(lr){11-12}
&& \multicolumn{2}{c}{MNIST} & \multicolumn{2}{c}{Pneumonia} & \multicolumn{2}{c}{Blood} & \multicolumn{2}{c}{Pneumonia}& \multicolumn{2}{c}{VinDr-CXR} \\
& $\#$Par. &ACC$\uparrow$ & AUC$\uparrow$ & ACC$\uparrow$ & AUC$\uparrow$ &  SSIM$\uparrow$ & PSNR$\uparrow$ & SSIM$\uparrow$ & PSNR$\uparrow$ &  ACC$\uparrow$ & IoU$\uparrow$ \\
 FFA &2.8m&\textbf{78.2} &\textbf{87.9} & \textbf{84.9} & \textbf{89.5} & $\underline{48.3}$ & $\underline{16.7}$ & $\textbf{72.3}$ & $\underline{16.8}$ & \textbf{93.2} & \textbf{40.1}\\
 CFFA &190k& 72.3 & $\underline{86.5}$ & 84.0 & 88.7 & $\underline{48.3}$ & $\underline{16.7}$ &  $\underline{71.6}$ & $\textbf{17.3}$ & 90.1 & 24.7\\
 ResCFFA &150k& $\underline{73.4}$ & 85.2 & $\underline{84.5}$ & $\underline{89.2}$ & \textbf{50.1} & \textbf{17.1} & 63.8 & 16.3 & $\underline{91.4}$ & $\underline{25.8}$\\
 \label{tab:pre}
\end{tabular}
}
\vspace{-1cm}
\end{table}

\section{Conclusion}
We presented a fast Self-adapting Forward-Forward Network (\emph{SaFF-Net}) for medical imaging analysis, mitigating power consumption and resource limitations without reliance on back-propagation for model training and fine-tuning. SaFF-Net shows remarkable performance with a notable $70$\% reduction in NFEs and $35$\% in parameters.

\noindent\emph{Acknowledgements}:  The authors gratefully acknowledge the scientific support and HPC resources provided by the Erlangen National High Performance Computing Center (NHR@FAU) of the Friedrich-Alexander-Universität Erlangen-Nürnberg (FAU) under the NHR projects b143dc and b180dc. NHR funding is provided by federal and Bavarian state authorities. NHR@FAU hardware is partially funded by the German Research Foundation (DFG) – 440719683. Additional support was also received by the ERC - project MIA-NORMAL 101083647,  DFG KA 5801/2-1, INST 90/1351-1 and by the state of Bavaria.
%
%
%
\newpage
\bibliographystyle{splncs04}
\bibliography{paper}

\begin{thebibliography}{10}
\providecommand{\url}[1]{\texttt{#1}}
\providecommand{\urlprefix}{URL }
\providecommand{\doi}[1]{https://doi.org/#1}

\bibitem{ahamed2023ffcl}
Ahamed, M.A., Chen, J., Imran, A.A.Z.: Ffcl: Forward-forward contrastive learning for improved medical image classification. In: Medical Imaging with Deep Learning, short paper track (2023)

\bibitem{balntas2016learning}
Balntas, V., Riba, E., Ponsa, D., Mikolajczyk, K.: Learning local feature descriptors with triplets and shallow convolutional neural networks. In: Bmvc. vol.~1, p.~3 (2016)

\bibitem{bianchi2020reservoir}
Bianchi, F.M., Scardapane, S., L{\o}kse, S., Jenssen, R.: Reservoir computing approaches for representation and classification of multivariate time series. IEEE transactions on neural networks and learning systems  \textbf{32}(5),  2169--2179 (2020)

\bibitem{blalock2020state}
Blalock, D., Gonzalez~Ortiz, J.J., Frankle, J., Guttag, J.: What is the state of neural network pruning? Proceedings of machine learning and systems  \textbf{2},  129--146 (2020)

\bibitem{brenig2023study}
Brenig, J., Timofte, R.: A study of forward-forward algorithm for self-supervised learning. arXiv preprint arXiv:2309.11955  (2023)

\bibitem{NECO_a_00052}
Cireşan, D.C., Meier, U., Gambardella, L.M., Schmidhuber, J.: {Deep, Big, Simple Neural Nets for Handwritten Digit Recognition}. Neural Computation  \textbf{22}(12),  3207--3220 (12 2010). \doi{10.1162/NECO_a_00052}, \url{https://doi.org/10.1162/NECO\_a\_00052}

\bibitem{de2023mu}
De~Vita, F., Nawaiseh, R.M., Bruneo, D., Tomaselli, V., Lattuada, M., Falchetto, M.: $\mu$-ff: On-device forward-forward training algorithm for microcontrollers. In: 2023 IEEE International Conference on Smart Computing (SMARTCOMP). pp. 49--56. IEEE (2023)

\bibitem{deng2012mnist}
Deng, L.: The mnist database of handwritten digit images for machine learning research. IEEE Signal Processing Magazine  \textbf{29}(6),  141--142 (2012)

\bibitem{gandhi2023extending}
Gandhi, S., Gala, R., Kornberg, J., Sridhar, A.: Extending the forward forward algorithm. arXiv preprint arXiv:2307.04205  (2023)

\bibitem{giampaolo2023investigating}
Giampaolo, F., Izzo, S., Prezioso, E., Piccialli, F.: Investigating random variations of the forward-forward algorithm for training neural networks. In: 2023 International Joint Conference on Neural Networks (IJCNN). pp.~1--7. IEEE (2023)

\bibitem{gutmann2010noise}
Gutmann, M., Hyv{\"a}rinen, A.: Noise-contrastive estimation: A new estimation principle for unnormalized statistical models. In: Proceedings of the thirteenth international conference on artificial intelligence and statistics. pp. 297--304. JMLR Workshop and Conference Proceedings (2010)

\bibitem{he2016deep}
He, K., Zhang, X., Ren, S., Sun, J.: Deep residual learning for image recognition. In: Proceedings of the IEEE conference on computer vision and pattern recognition. pp. 770--778 (2016)

\bibitem{hinton2022forward}
Hinton, G.: The forward-forward algorithm: Some preliminary investigations. arXiv preprint arXiv:2212.13345  (2022)

\bibitem{hinton1986learning}
Hinton, G.E., Sejnowski, T.J., et~al.: Learning and relearning in boltzmann machines. Parallel distributed processing: Explorations in the microstructure of cognition  \textbf{1}(282-317), ~2 (1986)

\bibitem{hopwood2023one}
Hopwood, M.: One-class systems seamlessly fit in the forward-forward algorithm. arXiv preprint arXiv:2306.15188  (2023)

\bibitem{lagani2022comparing}
Lagani, G., Falchi, F., Gennaro, C., Amato, G.: Comparing the performance of hebbian against backpropagation learning using convolutional neural networks. Neural Computing and Applications  \textbf{34}(8),  6503--6519 (2022)

\bibitem{licua2023sleep}
Lic{\u{a}}, M.T., Dinucu-Jianu, D.: Sleep patterns in the forward-forward algorithm  (2023), \url{https://openreview.net/forum?id=q_lJooPbX_}

\bibitem{lillicrap2016random}
Lillicrap, T.P., Cownden, D., Tweed, D.B., Akerman, C.J.: Random synaptic feedback weights support error backpropagation for deep learning. Nature communications  \textbf{7}(1),  13276 (2016)

\bibitem{lorberbom2024layer}
Lorberbom, G., Gat, I., Adi, Y., Schwing, A., Hazan, T.: Layer collaboration in the forward-forward algorithm. In: Proceedings of the AAAI Conference on Artificial Intelligence. vol.~38, pp. 14141--14148 (2024)

\bibitem{nguyen2020vindrcxr}
Nguyen, H.Q., Lam, K., Le, L.T., Pham, H.H., Tran, D.Q., Nguyen, D.B., Le, D.D., Pham, C.M., Tong, H.T.T., Dinh, D.H., Do, C.D., Doan, L.T., Nguyen, C.N., Nguyen, B.T., Nguyen, Q.V., Hoang, A.D., Phan, H.N., Nguyen, A.T., Ho, P.H., Ngo, D.T., Nguyen, N.T., Nguyen, N.T., Dao, M., Vu, V.: Vindr-cxr: An open dataset of chest x-rays with radiologist's annotations (2020)

\bibitem{ororbia2023learning}
Ororbia, A.: Learning spiking neural systems with the event-driven forward-forward process. arXiv preprint arXiv:2303.18187  (2023)

\bibitem{ororbia2023predictive}
Ororbia, A., Mali, A.: The predictive forward-forward algorithm. arXiv preprint arXiv:2301.01452  (2023)

\bibitem{paliotta2023graph}
Paliotta, D., Alain, M., M{\'a}t{\'e}, B., Fleuret, F.: Graph neural networks go forward-forward. arXiv preprint arXiv:2302.05282  (2023)

\bibitem{papachristodoulou2023convolutional}
Papachristodoulou, A., Kyrkou, C., Timotheou, S., Theocharides, T.: Convolutional channel-wise competitive learning for the forward-forward algorithm. arXiv preprint arXiv:2312.12668  (2023)

\bibitem{pardientropy}
Pardi, M., Tortorella, D., Micheli, A.: Entropy based regularization improves performance in the forward-forward algorithm

\bibitem{salimans2017evolution}
Salimans, T., Ho, J., Chen, X., Sidor, S., Sutskever, I.: Evolution strategies as a scalable alternative to reinforcement learning. arXiv preprint arXiv:1703.03864  (2017)

\bibitem{spratling2017review}
Spratling, M.W.: A review of predictive coding algorithms. Brain and cognition  \textbf{112},  92--97 (2017)

\bibitem{stanley2002evolving}
Stanley, K.O., Miikkulainen, R.: Evolving neural networks through augmenting topologies. Evolutionary computation  \textbf{10}(2),  99--127 (2002)

\bibitem{tosato2023emergent}
Tosato, N., Basile, L., Ballarin, E., de~Alteriis, G., Cazzaniga, A., Ansuini, A.: Emergent representations in networks trained with the forward-forward algorithm. arXiv preprint arXiv:2305.18353  (2023)

\bibitem{wang2016low}
Wang, Y., Xia, L., Tang, T., Li, B., Yao, S., Cheng, M., Yang, H.: Low power convolutional neural networks on a chip. In: 2016 IEEE International Symposium on Circuits and Systems (ISCAS). pp. 129--132. IEEE (2016)

\bibitem{yamazaki2022spiking}
Yamazaki, K., Vo-Ho, V.K., Bulsara, D., Le, N.: Spiking neural networks and their applications: A review. Brain Sciences  \textbf{12}(7), ~863 (2022)

\bibitem{medmnistv1}
Yang, J., Shi, R., Ni, B.: Medmnist classification decathlon: A lightweight automl benchmark for medical image analysis. In: IEEE 18th International Symposium on Biomedical Imaging (ISBI). pp. 191--195 (2021)

\bibitem{medmnistv2}
Yang, J., Shi, R., Wei, D., Liu, Z., Zhao, L., Ke, B., Pfister, H., Ni, B.: Medmnist v2-a large-scale lightweight benchmark for 2d and 3d biomedical image classification. Scientific Data  \textbf{10}(1), ~41 (2023)

\bibitem{yang2023theory}
Yang, Y.: A theory for the sparsity emerged in the forward forward algorithm. arXiv preprint arXiv:2311.05667  (2023)

\bibitem{2304-10054}
Zheng, Z., Jia, X.: Complex mixer for medmnist classification decathlon. CoRR  \textbf{abs/2304.10054} (2023), \url{https://doi.org/10.48550/arXiv.2304.10054}

\end{thebibliography}
\newpage

\appendix
\begin{table}[t]
\begin{minipage}[t]{1.0\linewidth} 
\caption{Ablation Study for CFFA on MNIST and MedMNIST (Supervised). The CFFA network consists of 4 layers with [32,32,128,128] channels each (194,432 \# Params). Unless otherwise noted, networks are trained max. 10 epochs with 10,000 layer iterations for CFFA(\emph{One-shot}) and 1,000 layer iterations for CFFA. NFE - Forward and Backward passes $\times1e^{3}$, ES - Early Stopping}
\begin{tabularx}{\textwidth}{lcccccccccccccccc}
 & \multicolumn{6}{c}{MNIST} & \multicolumn{6}{c}{MedMNIST Pneumonia}\\
  \cmidrule(lr){2-7} \cmidrule(lr){8-13}
 &  \multicolumn{3}{c}{CFFA (\emph{One-shot})}  &  \multicolumn{3}{c}{CFFA}  &  \multicolumn{3}{c}{CFFA (\emph{One-shot})} &  \multicolumn{3}{c}{CFFA}\\
 \cmidrule(lr){2-4} \cmidrule(lr){5-7} \cmidrule(lr){8-10} \cmidrule(lr){11-13} 
 & ACC & AUC & NFE &  ACC & AUC & NFE  &  ACC & AUC & NFE &  ACC & AUC & NFE  \\
 \cmidrule(lr){2-2}\cmidrule(lr){3-3}\cmidrule(lr){4-4}\cmidrule(lr){5-5}\cmidrule(lr){6-6}\cmidrule(lr){7-7}\cmidrule(lr){8-8}\cmidrule(lr){9-9}\cmidrule(lr){10-10}\cmidrule(lr){11-11}\cmidrule(lr){12-12}\cmidrule(lr){13-13}\cmidrule(lr){14-14}
 Baseline & \textbf{0.94} & \textbf{0.99} & 80.0 & 0.86 & \underline{0.98} & \underline{80.0} & \underline{0.86} & 0.90 & 80.0 & 0.85 & \underline{0.90} & 80.0\\
 \midrule
 ES Iterations & 0.87 & \textbf{0.99} & \textbf{65.2} &  0.86 & \underline{0.98} & \underline{80.0} & \textbf{0.87} & \underline{0.92} & \textbf{20.0}  & 0.83 & 0.88 & \textbf{14.1}\\
 ES Layers & \textbf{0.94} & \textbf{0.99} & 80.0 & 0.87 & 0.96 & \textbf{66.0} & \underline{0.86} & 0.90 & 80.0 & 0.82 & 0.86 & \underline{35.0} \\
 ES Epoch & - & - & - & 0.86 & \underline{0.98} & \underline{80.0} & - & - & - & 0.85 & 0.89 & 80.0\\
 \midrule
 Pruning (0.05) & 0.91 & \textbf{0.99} & 80.0 & 0.85 & \textbf{0.99} & \underline{80.0} & 0.84 & 0.89 & 80.0 & 0.85 & \underline{0.90} & 80.0\\
 w/ Retraining & \underline{0.93} & \textbf{0.99} & 80.8 & 0.85 & \textbf{0.99} & 80.8 & 0.76 & 0.81 & 80.4 & 0.85 & \underline{0.90} & 80.8 \\
 \midrule
 Train. Thresh.  & 0.92 & \textbf{0.99} & 80.0 & \underline{0.89} & \textbf{0.99} & \underline{80.0} & \textbf{0.87} & 0.91 & 80.0 & \textbf{0.87} & \textbf{0.91} & 80.0\\
 Batch-Norm & 0.87 & \underline{0.98} & 80.0 & 0.86 & \underline{0.98} & \underline{80.0} & 0.85 & 0.89 & 80.0 & \underline{0.86} & 0.89 & 80.0 \\
 \midrule
\emph{SaFF-Net} & 0.90 & \textbf{0.99} & \underline{71.8} & \textbf{0.90} & \textbf{0.99} & \underline{80.0} & \textbf{0.87} & \textbf{0.93} & \underline{37.9} & \textbf{0.87} & \textbf{0.91} & 80.8\\
\\
 \label{tab:abl-cffa}
\end{tabularx}
\end{minipage}
\begin{minipage}[t]{1.0\linewidth} 
\centering
\includegraphics[width=1.05\linewidth]{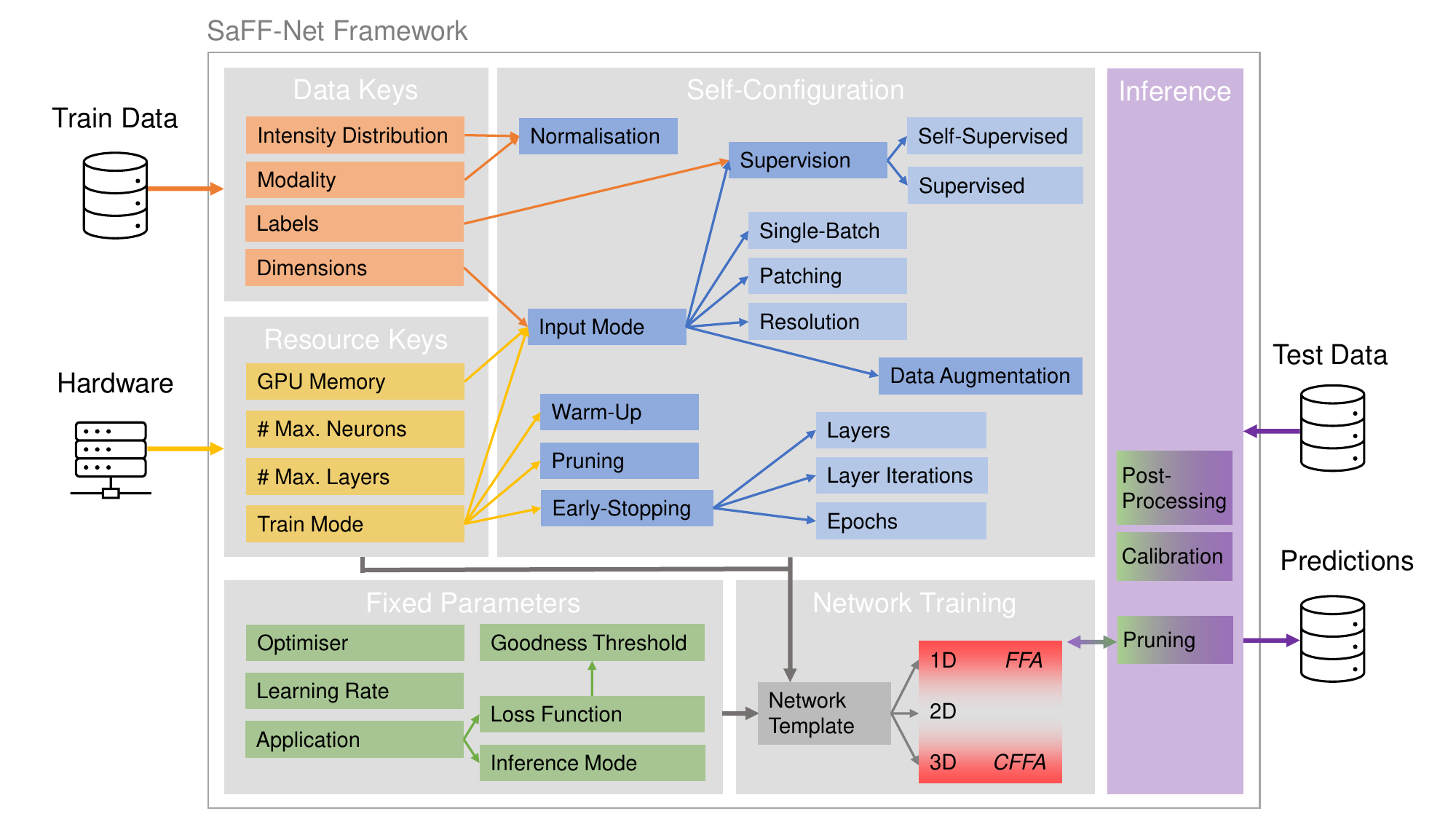}
\captionof{figure}{Self-adapting \emph{SaFF-Net} framework. Key features of the training data and the hardware components are used for self-configuration. 
The fixed parameters for the pipeline are given by default or via an experiment file. After self-configuration, the \emph{SaFF-Net}  selects the best network configuration and starts training. Inference, postprocessing, calibration and pruning can be enabled.}
\label{fig:scheme}
\end{minipage}
\end{table}

\begin{table}[]
\caption{Supervised Classification (Batchsize 60,000/MAX; Early Stopping, max 100 Epochs, Seed). \emph{SaFF-Net} with FFA is used for 1D (flatten) input data and CFFA is used for 2D/3D input data. Bin. CXR - Binary Classification (Cardiomegaly) on VinDr-CXR, ACC - Accuracy, AUC - Area Under the Receiver Operating characteristic, AP - Mean Average Precision.}
\vspace{6cm}
\rotatebox{90}{
\begin{tabularx}{\textwidth}{llcccccccccccccccccccccc}
 &  &   &   &   &  &  &  &   &  &  & \multicolumn{9}{c}{SaFF-Net \emph{(Ours)}} \\
 \cmidrule(lr){12-20}
 &  &  \multicolumn{3}{c}{MLP}  &  \multicolumn{3}{c}{CNN}  &  \multicolumn{3}{c}{ResNet18} &  \multicolumn{3}{c}{FFA} &  \multicolumn{3}{c}{CFFA} &  \multicolumn{3}{c}{ResCFFA}\\
 \cmidrule(lr){3-5} \cmidrule(lr){6-8} \cmidrule(lr){9-11} \cmidrule(lr){12-14} \cmidrule(lr){15-17} \cmidrule(lr){18-20}
 &  & ACC & AUC & AP &  ACC & AUC & AP  &  ACC & AUC & AP &  ACC & AUC & AP &  ACC & AUC & AP &  ACC & AUC & AP   \\
 \cmidrule(lr){3-3}\cmidrule(lr){4-4}\cmidrule(lr){5-5}\cmidrule(lr){6-6}\cmidrule(lr){7-7}\cmidrule(lr){8-8}\cmidrule(lr){9-9}\cmidrule(lr){10-10}\cmidrule(lr){11-11}\cmidrule(lr){12-12}\cmidrule(lr){13-13}\cmidrule(lr){14-14}\cmidrule(lr){15-15}\cmidrule(lr){16-16}\cmidrule(lr){17-17}\cmidrule(lr){18-18}\cmidrule(lr){19-19}\cmidrule(lr){20-20}
 & MNIST & 0.798 &  0.959 & 0.796 & 0.512 & 0.859 & 0.730 & 0.921 & 0.990 & 0.961 & 0.970  & 0.999 & 0.993 & 0.873 & 0.987 & 0.916 & 0.928 & 0.986 & 0.993 \\
 \cmidrule{1-20}
 \multirow{12}{*}{\rotatebox{90}{\makecell[l]{MedMNIST}}} & Path & 0.414 & 0.787 & 0.359 & 0.405 & 0.771 & 0.428 & 0.623 & 0.922 & 0.599 & 0.451 & 0.808 & 0.407 &  0.463 & 0.712  & 0.278 & 0.410 & 0.837 & 0.407\\
 & Chest & 0.638 & 0.883 & 0.085 & 0.640 & 0.886 & 0.082 & 0.632 & 0.891 &  0.094 & 0.429 & 0.738 & 0.097 & 0.425 & 0.726 & 0.085 & 0.256 & 0.566 & 0.097\\
 & Derma & 0.695 & 0.897 & 0.195 & 0.695 & 0.904 & 0.254 & 0.743 & 0.933 & 0.332 & 0.620 & 0.839 & 0.346 & 0.236 & 0.558 & 0.184 & 0.253 & 0.735 & 0.253\\
 & OCT & 0.356 & 0.546 & 0.348 & 0.250 & 0.537 & 0.342 & 0.415 & 0.634 & 0.451 & 0.626 & 0.860 & 0.663 & 0.702 & 0.896 & 0.677 & 0.667 & 0.893 & 0.723 \\
 & Pneum.  & 0.677 & 0.828 & 0.902 & 0.633 & 0.472 & 0.345 & 0.789 & 0.846 & 0.811 & 0.886 &  0.912 & 0.886 & 0.870 & 0.900 & 0.840 & 0.851 & 0.930 & 0.914\\
 & Retina & 0.435 & 0.680 & 0.274 & 0.435 & 0.756 & 0.310 & 0.493 & 0.737 & 0.322 & 0.463 & 0.733 & 0.354 & 0.382 & 0.661 & 0.313 & 0.405 & 0.733 & 0.388\\
 & Breast & 0.731 & 0.820 & 0.677 &  0.731 & 0.773 & 0.577 & 0.859 & 0.887 & 0.805 & 0.750 & 0.809 & 0.762 & 0.743 & 0.784 & 0.674 & 0.686 & 0.748 & 0.730\\
 & Blood & 0.384 & 0.837 & 0.408 & 0.380 & 0.792 & 0.315 & 0.805 & 0.971 & 0.805 & 0.630 & 0.907 & 0.671 & 0.532 & 0.800 & 0.511 & 0.556 & 0.919 & 0.706 \\
 & Tissue & 0.093  & 0.577 & 0.170 & 0.242 & 0.698 & 0.229 & 0.483 & 0.826 & 0.312 & 0.437 & 0.830 & 0.372 & 0.330 & 0.739 & 0.267 & 0.341 & 0.692 & 0.197 \\
 & OrganA & 0.428 & 0.790 & 0.454 & 0.194 & 0.745 & 0.341 & 0.753 & 0.962 & 0.799 & 0.589 & 0.878 & 0.515 & 0.366 & 0.771 & 0.277 & 0.401 & 0.860 & 0.492\\
 & OrganC & 0.397 & 0.775 & 0.408 & 0.222 & 0.676 & 0.322 & 0.747 & 0.957 & 0.785 & 0.636 &  0.898 & 0.570 & 0.332 & 0.687 & 0.234 & 0.390 & 0.856 & 0.481\\
 & OrganS & 0.179 & 0.721 & 0.282 & 0.236 & 0.687 & 0.224 & 0.579 & 0.918 & 0.553 & 0.452 & 0.848 & 0.412 & 0.251 & 0.633 & 0.198 & 0.278 & 0.766 & 0.354\\
 \cmidrule{1-20}
 \multirow{6}{*}{\rotatebox{90}{\makecell[l]{MedMNIST 3D}}}& Organ & 0.213 &  0.665 & 0.474 & 0.138 & 0.626 & 0.163 & 0.615 & 0.913 & 0.652 & 0.232 & 0.674 & 0.270 & 0.195 & 0.605 & 0.145 & 0.281 & 0.762 & 0.273 \\
 & Nodule & 0.794 & 0.834 & 0.639 & 0.794 & 0.758 & 0.465 & 0.809 & 0.841 & 0.606 & 0.790 & 0.819 & 0.660 & 0.553 & 0.576 & 0.575 & 0.573 & 0.635 & 0.609 \\
 & Adrenal & 0.768 & 0.827 & 0.604 & 0.768 & 0.758 & 0.480 & 0.768 & 0.776 & 0.523 & 0.631 & 0.587 & 0.576 & 0.737 & 0.765 & 0.564 & 0.683 & 0.671 & 0.591 \\
 & Fracture & 0.375 & 0.542 & 0.475 & 0.375 & 0.599 & 0.357 & 0.407 & 0.581 & 0.340 & 0.383  & 0.527 & 0.391 & 0.375 & 0.531 & 0.333 & 0.309 & 0.551 & 0.429\\
 & Vessel & 0.887 & 0.877 & 0.491 & 0.887 & 0.884 & 0.491 & 0.874 & 0.891 & 0.527 & 0.759 & 0.794 & 0.632 & 0.113 & 0.113 & 0.500 & 0.534 & 0.535 & 0.482\\
 & Synapse & 0.270 & 0.302 & 0.551 & 0.730 & 0.718 & 0.487 & 0.709 & 0.710 & 0.515 & 0.270  & 0.238 & 0.474 &  0.730 & 0.738 & 0.517 & 0.683 & 0.708 & 0.501\\
 \cmidrule{1-20}
  \multicolumn{2}{c}{Bin. CXR} & 0.999 & 0.999 & 0.500 & 0.986 & 0.978 & 0.500 & 0.898 & 0.895 & 0.500 & 0.899 & 0.961 & 0.500 & 0.999 & 0.999 & 0.500 & 0.999 & 0.999 & 0.500\\
 \cmidrule(lr){3-5} \cmidrule(lr){6-8} \cmidrule(lr){9-11} \cmidrule(lr){12-14} \cmidrule(lr){15-17} \cmidrule(lr){18-20}
   & \#Par. &  \multicolumn{3}{c}{2,797,010}  &  \multicolumn{3}{c}{1,554,954}  &  \multicolumn{3}{c}{11,183,694} &  \multicolumn{3}{c}{max. 2,797,010} &  \multicolumn{3}{c}{194,432} &  \multicolumn{3}{c}{148,933}\\
 \label{tab:full}
\end{tabularx}
}
\end{table}
\end{document}